%% file: emnlp2020.tex
\documentclass[11pt,a4paper]{article}
\usepackage[hyperref]{emnlp2020}
\usepackage{times}
\usepackage{latexsym}
\usepackage{amssymb}
\usepackage{pgfplots}
\usepackage{scalefnt}
\usepackage{boldline}
\usepackage{xcolor}
\usepackage{multirow}
\usepackage{makecell}
\usepackage{graphicx}
\usepackage{bm}
\usepackage{amsmath}
\usepackage{bbm}
\usepackage{float}
\usepackage{stfloats}
\usepackage{lipsum}
\usepackage{comment}
\usepackage{booktabs}
\usepackage{subfig}

\usepackage{microtype}

\aclfinalcopy 

\newcommand\blfootnote[1]{%
  \begingroup
  \renewcommand\thefootnote{}\footnote{#1}%
  \addtocounter{footnote}{-1}%
  \endgroup
}
\input{commands}

\title{Look at the First Sentence:\\Position Bias in Question Answering}

\author{Miyoung Ko \quad Jinhyuk Lee$^\dagger$ \quad Hyunjae Kim \quad Gangwoo Kim \quad Jaewoo Kang$^\dagger$\\
Korea University\\
\texttt{\{miyoungko,jinhyuk\_lee,hyunjae-kim\}@korea.ac.kr} \\
\texttt{\{gangwoo\_kim,kangj\}@korea.ac.kr}
}

\date{}

\begin{document}
\maketitle
\begin{abstract}
Many extractive question answering models are trained to predict start and end positions of answers.
The choice of predicting answers as positions is mainly due to its simplicity and effectiveness.
In this study, we hypothesize that when the distribution of the answer positions is highly skewed in the training set (e.g., answers lie only in the $k$-th sentence of each passage), QA models predicting answers as positions can learn spurious positional cues and fail to give answers in different positions.
We first illustrate this \textit{position bias} in popular extractive QA models such as BiDAF and BERT and thoroughly examine how position bias propagates through each layer of BERT.
To safely deliver position information without position bias, we train models with various de-biasing methods including entropy regularization and bias ensembling.
Among them, we found that using the prior distribution of answer positions as a bias model is very effective at reducing position bias, recovering the performance of BERT from 37.48\% to 81.64\% when trained on a biased SQuAD dataset.
\end{abstract}

\blfootnote{\textsuperscript{$\dagger$}Corresponding authors}

\section{Introduction}

\begin{figure} [t]
    \centering
    \includegraphics[width=0.9\columnwidth]{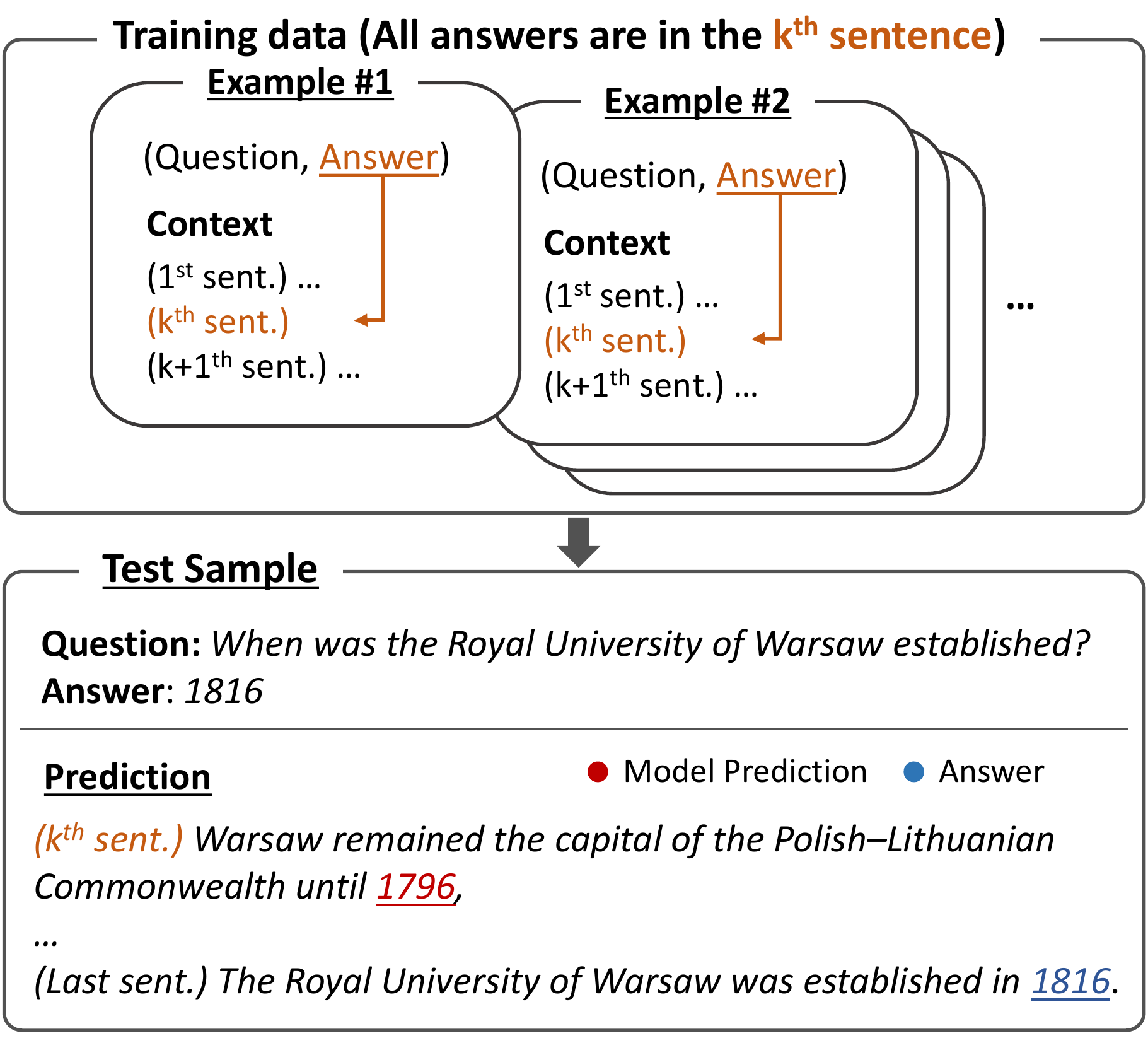}
    \vspace{-0.1cm}\caption{Example of position bias. BERT trained on the dataset with a skewed answer position distribution, provides wrong predictions, biased to the specific sentence position.}\vspace{-0.3cm}
    \label{fig:teaser}
\end{figure}

Question answering (QA) is a task of answering questions given a passage.
Large-scale QA datasets have attracted many researchers to build effective QA models, and with the advent of deep learning, recent QA models are known to outperform humans in some datasets~\citep{rajpurkar2016squad,devlin2019bert,yang2019xlnet}.
Extractive QA is the task that assumes that answers always lie in the passage. 
Based on this task assumption, various QA models are trained to predict the start and end positions as the answers.
Following the structure of earlier deep learning-based QA models~\citep{wang2016machine,seo2017bidirectional,xiong2017dcn+}, recent QA models provide positions of answers without much consideration~\citep{yu2018qanet,devlin2019bert,yang2019xlnet}.

The popularity of predicting the answer positions is credited to the fact that it reduces the prediction space to $\mathcal{O}(n)$ where $n$ is the length of an input document.
It is more efficient and effective than directly generating answers from a large vocabulary space.
Furthermore, it reduces the QA task to a classification task which is convenient to model.
Nevertheless, very few studies have discussed the side effects of predicting the answer positions.
Could there be any unwanted biases when using answer positions as prediction targets?

\begin{table*} [t]
\centering
{\setlength{\extrarowheight}{1pt}
\resizebox{\textwidth}{!}{
\def\arraystretch{0.9}
\begin{tabular}{l|ccc|ccc|ccc}
\toprule
\multirow{2}{*}{Training Data} & \multicolumn{3}{c|}{\textbf{BiDAF}} & \multicolumn{3}{c|}{\textbf{BERT}} &\multicolumn{3}{c}{\textbf{XLNet}} \\
& EM & F1 & $\Delta$ & EM & F1 & $\Delta$ & EM & F1 & $\Delta$ \\ \midrule
\textbf{\squad}\train               & 66.51 & 76.46 &  & 81.32 & 88.63 & & 80.69 & 89.24 &  \\
\textbf{\squad}\train ~(Sampled)    & 58.76 & 70.52 & -5.94 & 76.48 & 85.06  & -3.57 & 80.07 & 88.32  & -0.92  \\
\textbf{\squad}\trainone            & 21.44 & 27.92 & -48.54 & 31.20 & 37.48 & -51.15 & 38.59 & 45.27 & -43.97 \\
\textbf{\squad}\trainone~+ First Sentence       & 53.16 & 63.21  & -13.25 & 72.75 & 81.18  & -7.45 & 74.85 & 82.84  & -6.40 \\ 
\textbf{\squad}\trainone~+ Sentence Shuffle     & 54.40 & 65.20  & -11.26 & 73.37 & 81.90  & -6.73 & 77.83 & 86.18  & -3.06  \\
\bottomrule
\end{tabular}}}
\vspace{-0.1cm}\caption{Performance of QA models trained on the biased \squad~dataset (\squad\trainone), and tested on \squad\valid.
$\Delta$ denotes the difference in F1 score with~\squad\train.
We use exact match (EM) and F1 score for evaluation.\footnotemark}
\vspace{-0.4cm}\label{synthetic result}
\end{table*}

In this paper, we demonstrate that the models predicting the position can be severely biased when trained on datasets that have a very skewed answer position distribution.
We define this as position bias as shown in Figure~\ref{fig:teaser}.
Models trained on a biased dataset where answers always lie in the same sentence position mostly give predictions on the corresponding sentence.
As a result, BERT~\citep{devlin2019bert} trained on a biased training set where every answer appear in the first sentence only achieves 37.48\% F1 score in the SQuAD development set whereas the same model trained on the same amount of randomly sampled examples achieves 85.06\% F1 score.

To examine the cause of the problem, we thoroughly analyze the learning process of QA models trained on the biased training sets, especially focusing on BERT.
Our analysis shows that hidden representations of BERT preserve a different amount of word information according to the word position when trained on the biased training set.
The predictions of biased models also become more dependent on the first few words when the training set has answers only in the first sentences.

To tackle the problem, we test various options, ranging from relative position encodings~\citep{yang2019xlnet} to ensemble-based de-biasing methods~\citep{clark2019don, he2019unlearn}.
While simple baselines motivated by our analysis improve the test performance, our ensemble-based de-biasing method largely improves the performance of most models.
Specifically, we use the prior distribution of answer positions as an additional bias model and train models to learn reasoning ability beyond the positional cues.

Contributions of our paper are in threefold;
First, we define position bias in extractive question answering and illustrate that common extractive QA models suffer from it.
Second, we examine the reason for the failure of the biased models and show that positions can act as spurious biases.
Third, we show that the prior distribution of answer positions helps us to build positionally de-biased models, recovering the performance of BERT from 37.48\% to 81.64\%.
We also generalize our findings in many different positions and datasets.\footnote{\url{https://github.com/dmis-lab/position-bias}} 

\begin{figure*} [t]
    \centering
     \subfloat[Average cosine similarity (Layer 12)]
     {\includegraphics[width=0.55\textwidth]{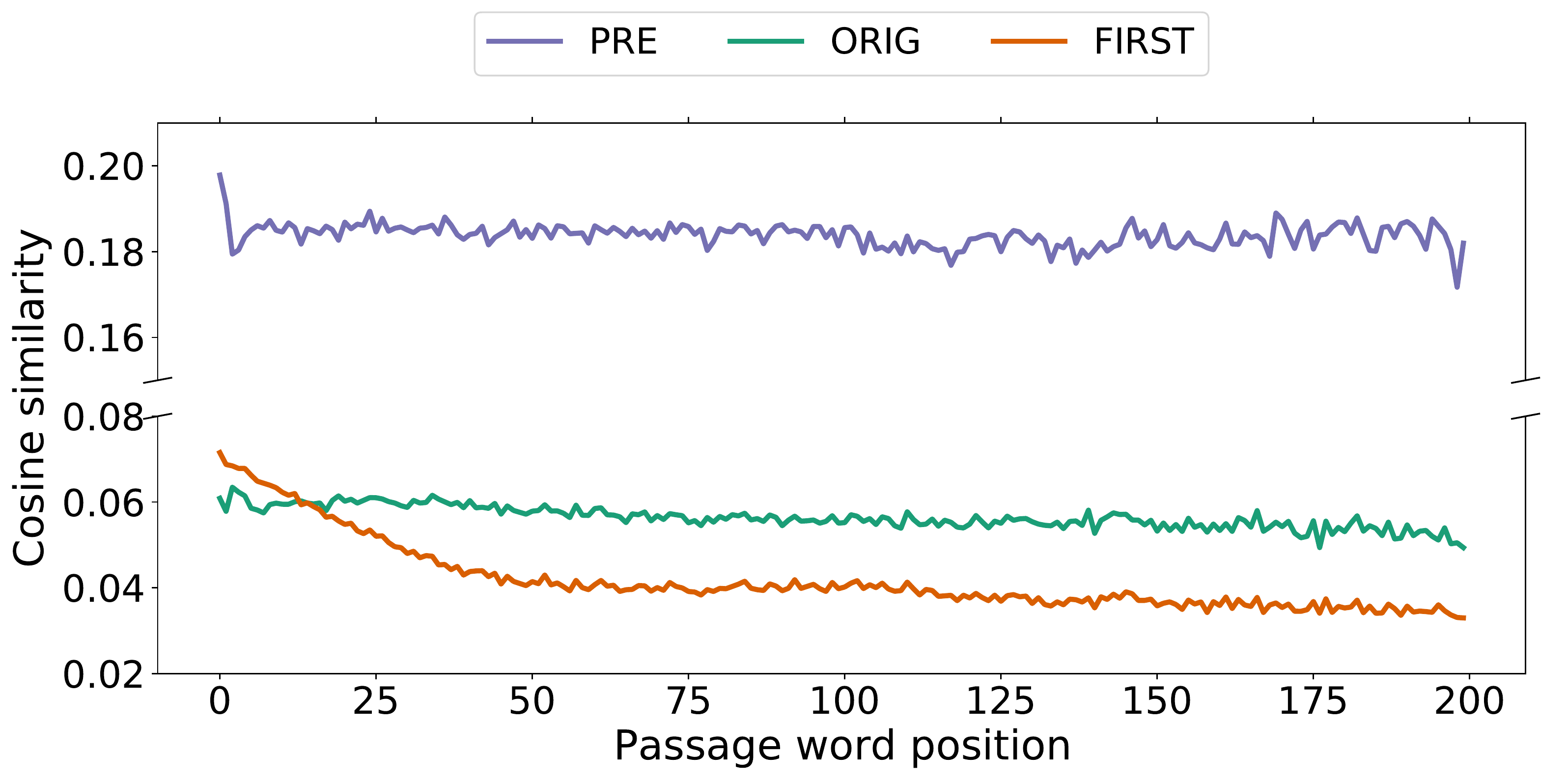}}
     \subfloat[Spearman correlation (start)]
     {\includegraphics[width=0.35\textwidth]{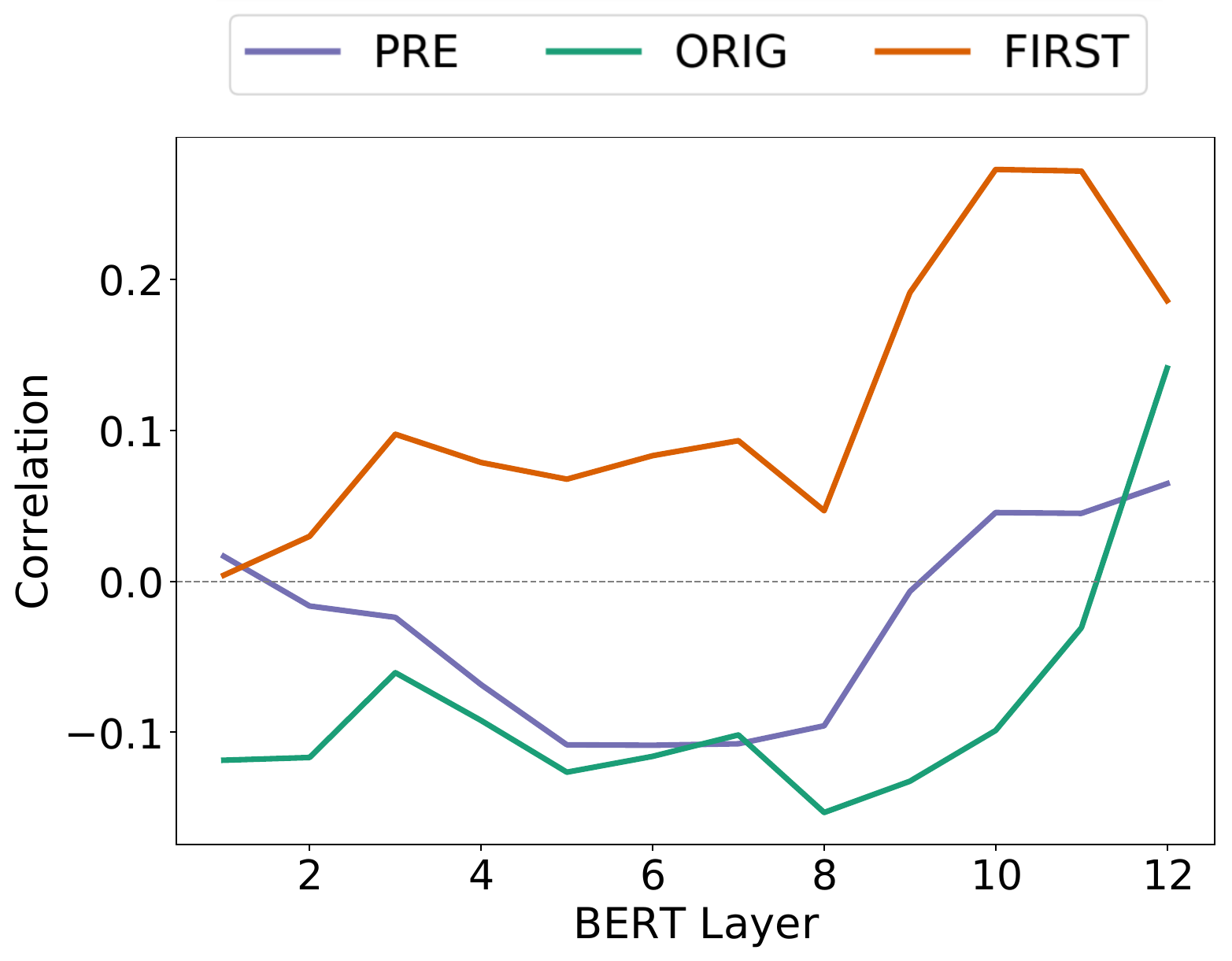}}
    \caption{Visualization of position bias with BERT trained on \squad\train~(\textcolor{green1}{\textsc{Orig}}), \squad\trainone~(\textcolor{orange}{\textsc{First}}), and BERT without fine-tuning (\textcolor{purple1}{\textsc{Pre}}). See Section  See Section~\ref{sec:2_2} for more details.}\vspace{-0.3cm}
    \label{fig:cos_sim_bias}
\end{figure*}

\footnotetext{Evaluation code is provided by \url{https://rajpurkar.github.io/SQuAD-explorer/}}

\section{Analysis}\label{sec:2}
We first demonstrate the presence of position bias using biased training sets sampled from SQuAD~\citep{rajpurkar2016squad} and visualize how position bias propagates in BERT.

\subsection{Position Bias on Synthetic Datasets}\label{section:2_1}
From the original training set \bigd\train, we subsample a biased training set \bigd\traink whose answers lie in the $k$-th sentence.\footnote{We use Spacy Sentencizer (\url{https://spacy.io/api/sentencizer}) for the sentence split.}
We conduct experiments on \squad~(\bigd~= SQuAD) as most examples in \squad~are answerable with a single sentence~\citep{min2018efficient}.
Our analysis mainly focuses on \squad\trainone~(i.e., all answers are in the first sentence), which has the largest proportion of samples compared to other sentence positions in \squad~(28,263 out of 87,599).
The proportion in the development set (\squad\valid) is similar, having 3,637 out of 10,570 answers in the first sentence.
Note that while our analysis is based on \squad\trainone, we also test various sentence positions in our main experiments (Section~\ref{sec:4.2}).
We experiment with three popular QA models that provide positions as answers: BiDAF~\citep{seo2017bidirectional}, BERT~\citep{devlin2019bert}, and XLNet~\citep{yang2019xlnet}.
All three models are trained on \squad\trainone~and are evaluated on \squad\valid.
For a fair comparison, we also randomly sample examples from the original training set and make \squad\train~(Sampled) which has the same number of examples with \squad\trainone.

Table~\ref{synthetic result} shows the performance of the three models trained on \squad\trainone. 
The performances of all models drop significantly compared to the models trained on \squad\train~or \squad\train~(Sampled).
The relative position encodings in XLNet mitigate position bias to some extent, but its performance still degrades significantly.

To better understand the cause of position bias, we additionally perform two pre-processing methods on \squad\trainone.
First, we truncate each passage up to the first sentence (\squad\trainone~+ First Sentence).
In this case, most performance is recovered, which indicates that the distributions of answer positions are relatively defined with respect to the maximum sequence length.
Shuffling the sentence order of \squad\trainone~
(\squad\trainone~+ Sentence Shuffle) also recovers most performance, showing that the spreadness of answers matters.
However, these pre-processing methods cannot be a solution as more fine-grained biases (e.g., word level positions) could cause the problem again and models cannot learn proper multi-sentence reasoning from a corrupted context.

\subsection{Visualization of Position Bias}\label{sec:2_2}
To visualize how position bias propagates throughout the layers, 
we compare BERT models, each trained on \squad\trainone~and \squad\train~respectively and BERT without any fine-tuning.
The uncased version of BERT-base is used for the analysis.

Figure~\ref{fig:cos_sim_bias} (a) shows the amount of word information preserved in the hidden representations at the last layer of BERT.
We define the amount of word information for each word position as the cosine similarity between the word embedding and its hidden representation at each layer.
The similarities are averaged over the passage-side hidden representations in \squad\valid.

BERT trained on \squad\trainone~(\textcolor{orange}{\textsc{First}}) has higher similarities at the front of the passages compared with BERT trained on \squad\train~(\textcolor{green1}{\textsc{Orig}}).
In the biased model, the similarity becomes smaller after the first few tokens, which shows position bias of BERT.

Figure~\ref{fig:cos_sim_bias} (b) shows the Spearman's rank correlation coefficient between the final output logits\footnote{We show the results with start position logits and the same pattern is observed with end position logits.} and the amount of word information at each layer defined by the cosine similarity.
A higher correlation means that the model is more dependent on the word information kept in that layer.
The correlation coefficient is much higher in the biased model (\textcolor{orange}{\textsc{First}}), especially in the last few layers.
Combined with the observation from Figure~\ref{fig:cos_sim_bias} (a), this indicates that the predictions of the biased model are heavily relying on the information of the first few words.

\subsection{Why is Position Bias Bad?}
Our analysis shows that it is very easy for neural QA models to exploit positional cues whenever possible.
While it is natural for neural models to learn the strong but spurious correlation present in the dataset~\citep{mccoy2019right,niven2019probing}, we argue that reading ability should be cultivated independent of such positional correlation.
Our study aims to learn proper reading ability even in extreme cases where all answers are in the $k$-th sentence.
Although exploiting the position distribution within the dataset could help the model improve performance on its corresponding test set, position bias should not be learned since we cannot guarantee realistic test environments to follow similar distribution.

\section{Method} \label{method}
To prevent models from learning a direct correlation between word positions and answers, we introduce simple remedies for BERT and a bias ensemble method with answer prior distributions that can be applied to any QA models.

\subsection{Baselines}

\paragraph{Randomized Position} 
To avoid learning the direct correlation between word positions and answers, we randomly perturb input positions.
We first randomly sample $t$ indices from a range of 1 to maximum sequence length of BERT.
We sample $t=384$ when the maximum sequence length is 512.
Then, we sort the indices in an ascending order to preserve the ordering of input words.
Perturbed indices then generate position embedding at each token position, which replaces the original position embedding.

\paragraph{Entropy Regularization}
Inspired by the observation in Section~\ref{sec:2_2}, we force our model to preserve a constant amount of word information regardless of the word positions.
Maximizing the entropy of normalized cosine similarity between the word embeddings and their hidden representations encourages models to maintain a uniform amount of information.
As the cosine similarities are not probabilities, we normalize them to be summed to 1.
We compute the entropy regularization term from the last layer and add it to the start/end prediction loss with a scaling factor $\lambda$.

\subsection{Bias Ensemble with Answer Prior}
Bias ensemble methods~\citep{clark2019don, he2019unlearn, mahabadi2020endtoend} combine the log probabilities from a pre-defined bias model and a target model to de-bias.
Ensembling makes the target model to learn different probabilities other than the bias probabilities.
In our case, we define the prior distribution of the answer positions as our bias model.
Specifically, we introduce the sentence-level answer prior and the word-level answer prior.

\paragraph{Bias Ensemble Method}
Given a passage and question pair, a model has to find the optimal start and end positions of the answer in the passage, denoted as $y_s$, $y_e$.
Typically, the model outputs two probability distributions $p^s$ and $p^e$ for the start and end positions.
As our method is applied in the same manner for both start and end predictions, we drop the superscript from $p^s$, $p^e$ and subscript from $y_s$, $y_e$ whenever possible.

For ensembling two different log probabilities from the bias model and the target model, we use a product of experts~\citep{hinton2002training}.
Using the product of experts, a probability at the $i$-th position is calculated as:
\begin{equation}\label{eq:bias_prod}
    \hat{p}_{i} =  softmax(\log(p_{i}) + \log(b_{i}))
\end{equation}
where $\log(p_i)$ is a log probability from the target model and $\log(b_{i})$ is a log probability from the bias model.
The ensembled probability $\hat{p}$ is used for the training.

To dynamically choose the amount of bias for each sample, \citet{clark2019don}~introduce a learned mixing ensemble with a trainable parameter.
Probabilities in the training phase are now defined as:
\begin{equation}\label{eq:mixin}
    \hat{p}_{i} =  softmax(\log(p_{i}) + g(X)\log(b_{i}))
\end{equation}
We use hidden representations before the softmax layer as $X$.
$g(X)$ then applies affine transformation on the representations to obtain a scalar value. 
Softplus activation followed by max pooling is used to obtain positive values.
As BiDAF has separate hidden representations for the start and end predictions, we separately define $g(X)$ for each start and end representation.

As models often learn to ignore the biases and make $g(X)$ to 0, \citet{clark2019don} suggest adding an entropy penalty term to the loss function. 
However, the entropy penalty did not make much difference in our case as $g(X)$ was already large enough.
Note that we only use $\log(b_i)$ during training, and the predictions are solely based on the predicted log probability $\log(p_i)$ from the model.

We define bias log probability as pre-calculated answer priors.
Using prior distributions in machine learning has a long history such as using class frequency in the class imbalance problem~\citep{domingos1999metacost, japkowicz2002class, zhou2006multi, huang2016learning}. 
In our case, the class prior corresponds to the prior distribution of answer positions.

\paragraph{Word-level Answer Prior} 
First, we consider the word-level answer prior.
Given the training set having $N$ examples having $N$ answers $\{y^{(1)}, y^{(2)}, ..., y^{(N)}\}$, we compute the word-level answer prior at position $i$ over the training set.
In this case, our bias log probability at $i$-th position is:
\begin{equation}
\label{equation:word-level-prior}
    \log(b_i) := \frac{1}{N}\sum_{j=1}^N \mathbf{1}[y^{(j)} = i]
\end{equation}
where we use the indicator function $\mathbf{1}[cond]$.
Bias log probabilities for the end position prediction are calculated in a similar manner.
Note that the word-level answer prior gives an equal bias distribution for each passage while the distribution is more fine-grained than the sentence-level prior described in the next section.

\paragraph{Sentence-level Answer Prior}
We also use the sentence-level answer prior which dynamically changes depending on the sentence boundaries of each sample.
First, we define a set of sentences $\{S_1^{(j)}, ..., S_L^{(j)}\}$ for the $j$-th training passage, where $L$ is the maximum number of sentence in whole training passages.
Then, the sentence-level answer prior of the $i$-th word position (for the start prediction) for the $j$-th sample, is derived from the frequency of answers appearing in the $l$-th sentence:
\begin{equation}
    \log(b_i^{(j)}) := \frac{1}{N}\sum_{k=1}^N\mathbf{1}[y^{(k)} \in S_l^{(k)}], \ \ i \in S_l^{(j)}
\end{equation}
Note that as boundaries of sentences in each sample are different, bias log probabilities should be defined in every sample.
Again, bias log probabilities for the end positions are calculated similarly.

It is very convenient to calculate the answer priors for any datasets.
For instance, on \bigd\trainone, we use the first sentence indicator as the sentence-level answer prior as all answers are in the first sentence.
More formally, the sentence-level answer prior for \bigd\trainone~is 1 for $l=1$, and 0 when $l > 1$:
\begin{equation}
    \log(b_i^{(j)}) := \begin{cases}
    1 & \quad i \in S_1^{(j)}, \\
    0 & \quad i \notin S_1^{(j)}
    \end{cases}\
\end{equation}
which is a special case of the sentence-level answer prior.
For general datasets where the distributions of answer positions are less skewed, the answer priors are more softly distributed. 
See Appendix~\ref{example_app} for a better understanding of the answer priors.

Both word-level and sentence-level answer priors are experimented with two bias ensemble methods: product of experts with bias (\textbf{Bias Product}, Equation~\ref{eq:bias_prod}) and learned mixing of two log probabilities (\textbf{Learned-Mixin}, Equation~\ref{eq:mixin}). 

\begin{table*} [h]
\centering
{\setlength{\extrarowheight}{2.05pt}
\resizebox{0.9\textwidth}{!}{
\def\arraystretch{0.9}
\begin{tabular}{c|l|cc|cc|cc}
\toprule
\multicolumn{2}{l|}{\multirow{2}{*}{De-biasing Method}} & \multicolumn{2}{c|}{\textbf{\squad}\validone} & \multicolumn{2}{c|}{\textbf{\squad}\validafter} & \multicolumn{2}{c}{\textbf{\squad}\valid} \\
\multicolumn{2}{c|}{} & EM & F1 & EM & F1 & EM & F1 \\ \midrule
\multicolumn{8}{c}{\textbf{BERT} trained on \textbf{\squad}\trainone} \\\midrule
\multirow{3}{*}{Baseline} & None & 77.07 & 85.81 & 7.14 & 12.12 & 31.20 & 37.48\\
& Random Position                & 69.95 & 80.73 & 27.32 & 34.63 & 41.99 & 50.49 \\
& Entropy Regularization                      & 77.40 & 86.17 & 10.50 & 15.72 & 33.52 & 39.96 \\ \midrule
\multirow{2}{*}{Word-Level} & Bias Product    & \textbf{78.61} & \textbf{87.08} & 7.85 & 12.88 & 32.19 & 38.41 \\
& Learned-Mixin                             & 78.17 & 86.56 & 8.55 & 13.43 & 32.51 & 38.59 \\ \midrule
\multirow{2}{*}{Sentence-Level} & Bias Product & 78.39 & 87.06 & 13.33 & 18.73 & 35.71 & 42.24 \\
& Learned-Mixin                             & 77.18 & 85.15 & \textbf{71.31} & \textbf{79.79} & \textbf{73.33} & \textbf{81.64} \\ \midrule
\multicolumn{8}{c}{\textbf{BiDAF} trained on \textbf{\squad}\trainone}\\ \midrule
Baseline & None           & 61.04 & 72.91 & 0.66 & 4.34 & 21.44 & 27.92 \\ \midrule
\multirow{2}{*}{Sentence-Level} & Bias Product  & \textbf{62.00} & \textbf{73.87} & 0.78 & 4.48 & 21.84 & 28.36 \\
& Learned-Mixin                 & 56.53 & 66.79 & \textbf{50.28} & \textbf{60.77} & \textbf{52.43} & \textbf{62.84} \\ \midrule
\multicolumn{8}{c}{\textbf{XLNet} trained on \textbf{\squad}\trainone} \\ \midrule
Baseline & None & 78.99 & 87.24 & 11.52 & 16.77 & 38.59 & 45.27 \\ \midrule
\multirow{2}{*}{Sentence-Level} & Bias Product & \textbf{79.24} & \textbf{87.88} & 33.28 & 39.93 & 49.09 & 56.43 \\
& Learned-Mixin     & 68.82 & 82.05 & \textbf{64.63} & \textbf{77.65} & \textbf{66.07} & \textbf{79.16} \\ \midrule
\multicolumn{8}{c}{\textbf{BERT} trained on \textbf{\squad}\train} \\ \midrule
Baseline & None & 81.55 & 88.68 & 81.21 & 88.61 & 81.32 & 88.63 \\ \midrule
\multirow{2}{*}{Sentence-Level} & Bias Product  & \textbf{81.88} & \textbf{88.87} & \textbf{81.29} & \textbf{88.87} & \textbf{81.49} & \textbf{88.87} \\
& Learned-Mixin & 81.58 & 88.38 & 80.87 & 88.47 & 81.12 & 88.44 \\ \bottomrule
\end{tabular}
}}
\caption{Results of applying de-biasing methods. Each model is evaluated on \squad\valid~and two subsets: \squad\validone~and \squad\validafter.}
\label{debias_performance}\vspace{-0.25cm}
\end{table*}

\section{Experiments}
We first experiment the effects of various de-biasing methods on three different QA models using both biased and full training sets.
Our next experiments generalize our findings in different sentence positions and different datasets such as NewsQA~\citep{trischler2017newsqa} and NaturalQuestions~\citep{naturalquestions}.

\subsection{Effect of De-biasing Methods}
We first train all three models (BiDAF, BERT, and XLNet) on \squad\trainone~with our de-biasing methods and evaluate them on \squad\valid~(original development set), \squad\validone, and \squad\validafter.
Note that \squad\validafter~is another subset of \squad\valid, whose answers do not appear in the first sentence, but in other sentences.
We also experiment with BERT trained on the full training set, \squad\train.

For all models, we use the same hyperparameters and training procedures as suggested in their original papers~\citep{seo2017bidirectional,devlin2019bert,yang2019xlnet}, except for batch sizes and training epochs (See Appendix~\ref{sec:train_detail}).
$\lambda$ for the entropy regularization is set to 5.
Most of our implementation is based on the PyTorch library.

\paragraph{Results with \squad\trainone}
The results of applying various de-biasing methods on three models with \squad\trainone~are in Table~\ref{debias_performance}.
Performance of all models without any de-biasing methods (denoted as `None') is very low on \squad\validafter, but fairly high on \squad\validone.
This means that their predictions are highly biased towards the first sentences.
In the case of BERT, F1 score on \squad\validone~is 85.81\%, while F1 score on \squad\validafter~is merely 12.12\%.
Our simple baseline approaches used in BERT improve the performance up to 34.63\% F1 score (Random Position) while the entropy regularization is not significantly effective. 

Bias ensemble methods using answer priors consistently improve the performance of all models.
The sentence-level answer prior works the best, which obtains a significant gain after applying the Learned-Mixin method.
We found that the coefficient $g(X)$ in Equation~\ref{eq:mixin} averages to 7.42. during training for BERT + Learned-Mixin, which demonstrates a need of proper balancing between the probabilities.
The word-level answer prior does not seem to provide strong position bias signals as its distribution is much softer than the sentence-level answer prior. 

\paragraph{Results with \squad\train}
The results of training BERT with our de-biasing methods on the full training set~\squad\train~are in the bottom of Table~\ref{debias_performance}.
Note that the answer prior is more softened than the answer prior used in \squad\trainone~as answers are now spread in all sentence positions.
While exploiting the positional distribution of the training set could be more helpful when evaluating on the development set that has a similar positional distribution, our method maintains the original performance.
It shows that our method works safely when the positional distribution doesn't change much.

\begin{figure}
    \centering
    \includegraphics[width=0.98\columnwidth]{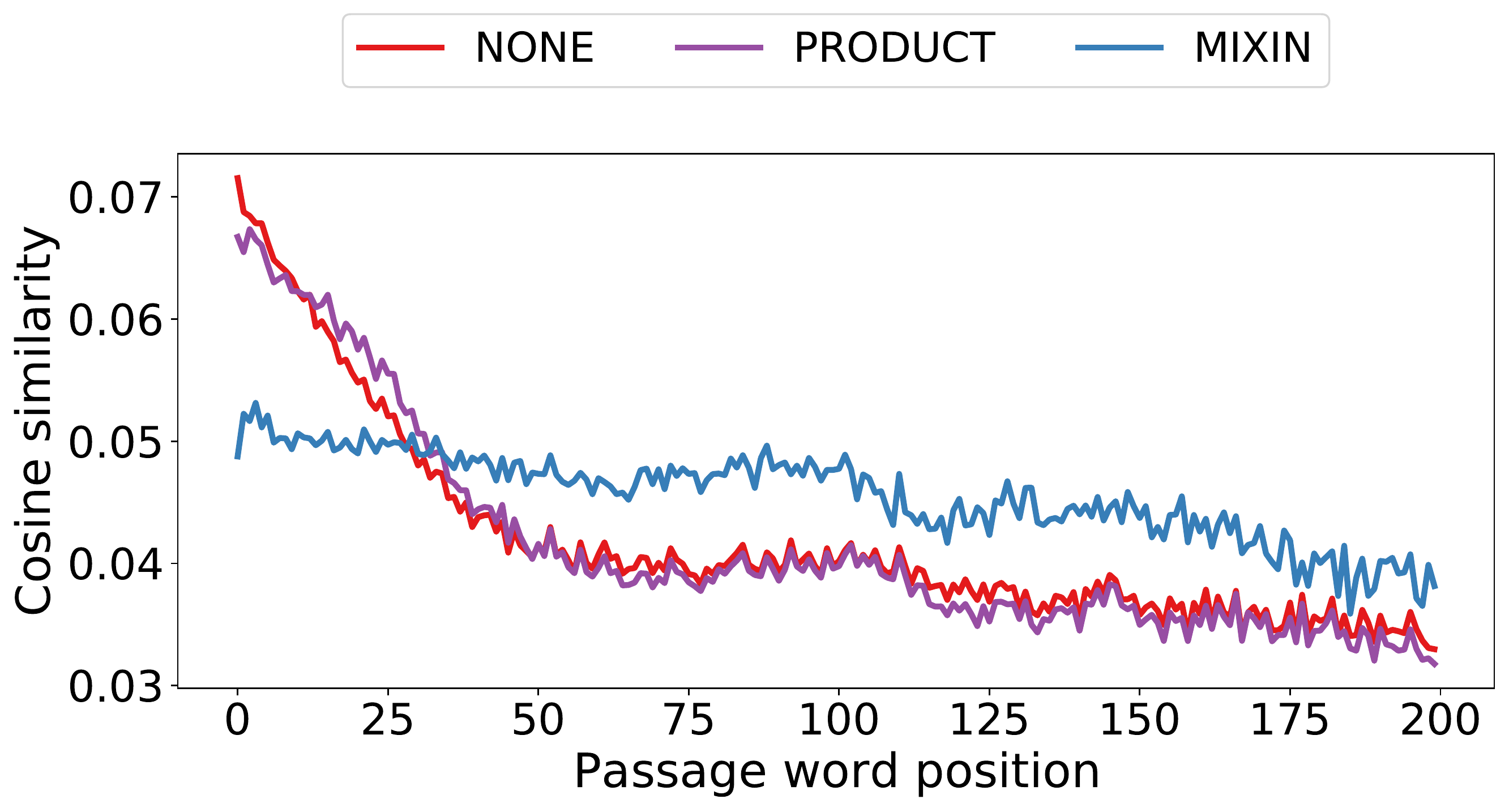}
    \caption{Visualization of BERT models trained on \squad\trainone~ with and without de-biasing method.}\vspace{-0.2cm}
    \label{fig:debias_trun}
\end{figure}

\paragraph{Visualization} \label{sec:4.1}
To investigate the effect of de-baising methods, we visualize the word information in each layer as done in Section~\ref{sec:2_2}.
We visualize the BERT trained on \squad\trainone~ensembled with sentence-level answer prior in Figure~\ref{fig:debias_trun}.
The bias product method (\textcolor{purple2}{\textsc{Product}}) and the model without any de-biasing methods (\textcolor{red}{\textsc{None}}) are similar, showing that it still has position bias.
The learned-mixin method (\textcolor{blue}{\textsc{Mixin}}), on the other hand, safely delivers the word information across different positions.

\subsection{Generalizing to Different Positions}  \label{sec:4.2}
As the \squad~training set has many answers in the first sentence, we mainly test our methods on \squad\trainone.
However, does our method generalize to different sentence positions?
To answer this question, we construct four \squad\traink datasets based on the sentence positions of answers.
Note that unlike \squad\trainone, the number of samples becomes smaller and the sentence boundaries are more blurry when $k > 1$, making answer priors much softer.
We train three QA models on different biased datasets and evaluate them on \squad\valid~with and without de-biasing methods.

\begin{table*} [t]
\centering\
\resizebox{0.9\textwidth}{!}{
\def\arraystretch{0.93}
\begin{tabular}{l|cc|cc|cc|cc}
\toprule
\multirow{2}{*}{} & \multicolumn{8}{c}{\textbf{\squad}\valid} \\
 & EM & F1 & EM & F1 & EM & F1 & EM & F1 \\\midrule
\multirow{2}{*}{\textbf{\squad}\traink} & \multicolumn{2}{c|}{$\bm{k=2}$} & \multicolumn{2}{c|}{$\bm{k=3}$} &\multicolumn{2}{c|}{$\bm{k=4}$} & \multicolumn{2}{c}{$\bm{k=5,6,...}$} \\
 & \multicolumn{2}{c|}{(20,593 samples)} & \multicolumn{2}{c|}{(15,567 samples)} &\multicolumn{2}{c|}{(10,379 samples)} & \multicolumn{2}{c}{(12,610 samples)} \\\midrule
\textbf{BiDAF}          & 18.43 & 25.74 & 12.26 & 19.04 & 9.96 & 16.50 & 12.34 & 19.65 \\
+Bias Product           & 21.51 & 28.67 & 11.19 & 18.39 & 11.20 & 17.78 & 10.09 & 16.78 \\
+Learned-Mixin          & \textbf{47.49} & \textbf{58.36} & \textbf{43.57} & \textbf{53.80} & \textbf{30.18} & \textbf{39.51} & \textbf{18.51} & \textbf{27.30} \\ \midrule
\textbf{BERT}           & 36.16 & 43.14 & 44.76 & 52.89 & 49.13 & 58.01 & 57.95 & 66.69  \\
+Bias Product           & 52.89 & 50.38 & 52.42 & 60.99 & 53.39 & 62.69 & 58.75 & 67.67 \\
+Learned-Mixin          & \textbf{71.61} & \textbf{80.36} & \textbf{69.04} & \textbf{77.91} & \textbf{64.31} & \textbf{73.72} & \textbf{62.82} & \textbf{72.30} \\ \midrule
\textbf{XLNet}          & 47.55 & 55.01 & 46.67 & 54.56 & 50.49 & 58.74 & 58.29 & 66.67  \\
+Bias Product           & 59.49 & 67.35 & 61.99 & 70.89 & 67.26 & 76.55 & 72.44 & 81.85 \\
+Learned-Mixin          & \textbf{68.34} & \textbf{80.35} & \textbf{69.28} & \textbf{79.99} & \textbf{70.07} & \textbf{80.12} & \textbf{73.33} & \textbf{82.79}  \\\bottomrule
\end{tabular}}
\caption{Position bias in different positions. Each model is trained on a biased SQuAD dataset (\squad\traink) and evaluated on \squad\valid.}\vspace{-0.1cm}
\label{other sentences}
\end{table*}

 \begin{figure*} [t]
    \centering
    \subfloat[BERT]
     {\includegraphics[width=0.32\textwidth]{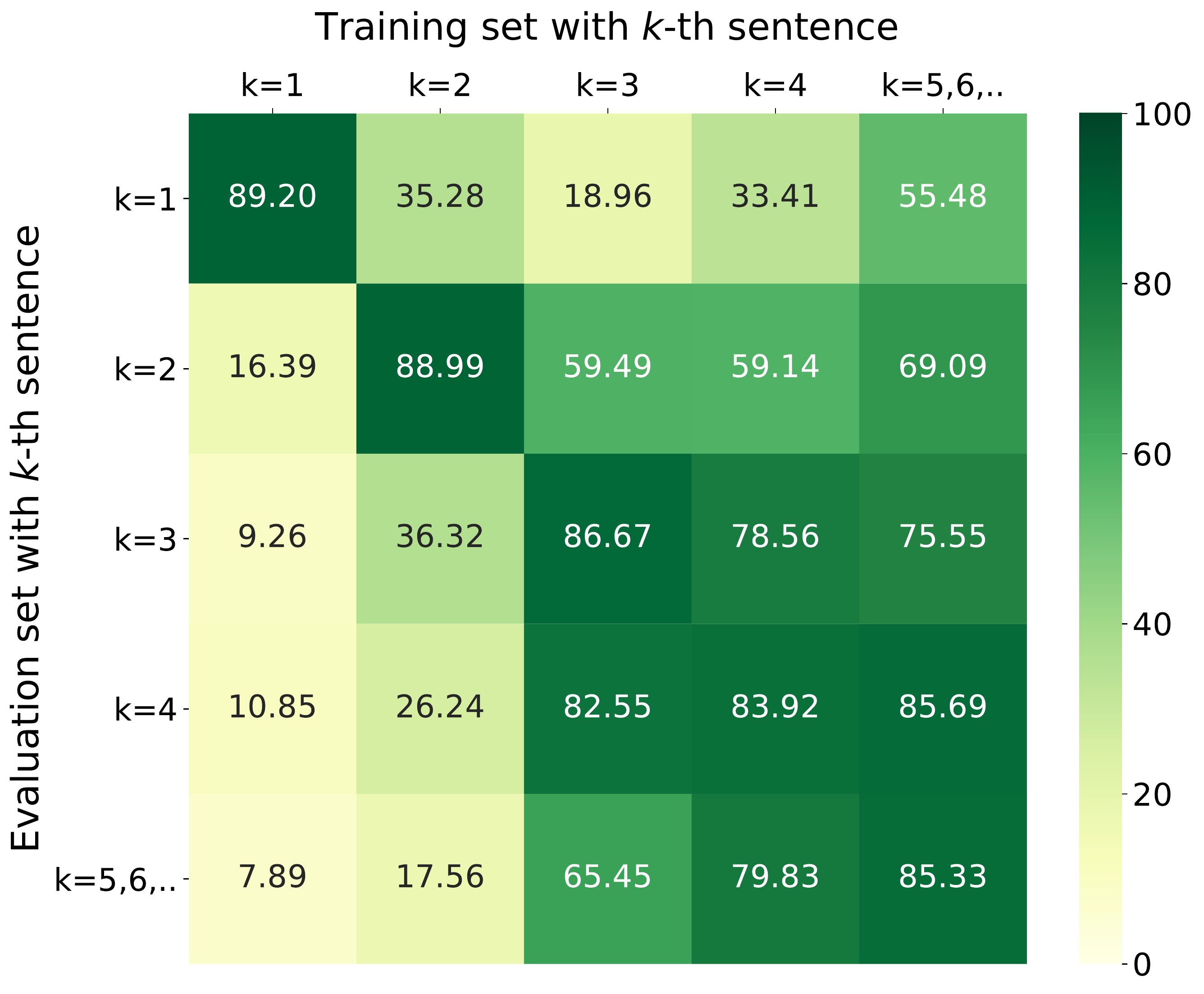}}
    \subfloat[BERT + Bias Product]
     {\includegraphics[width=0.32\textwidth]{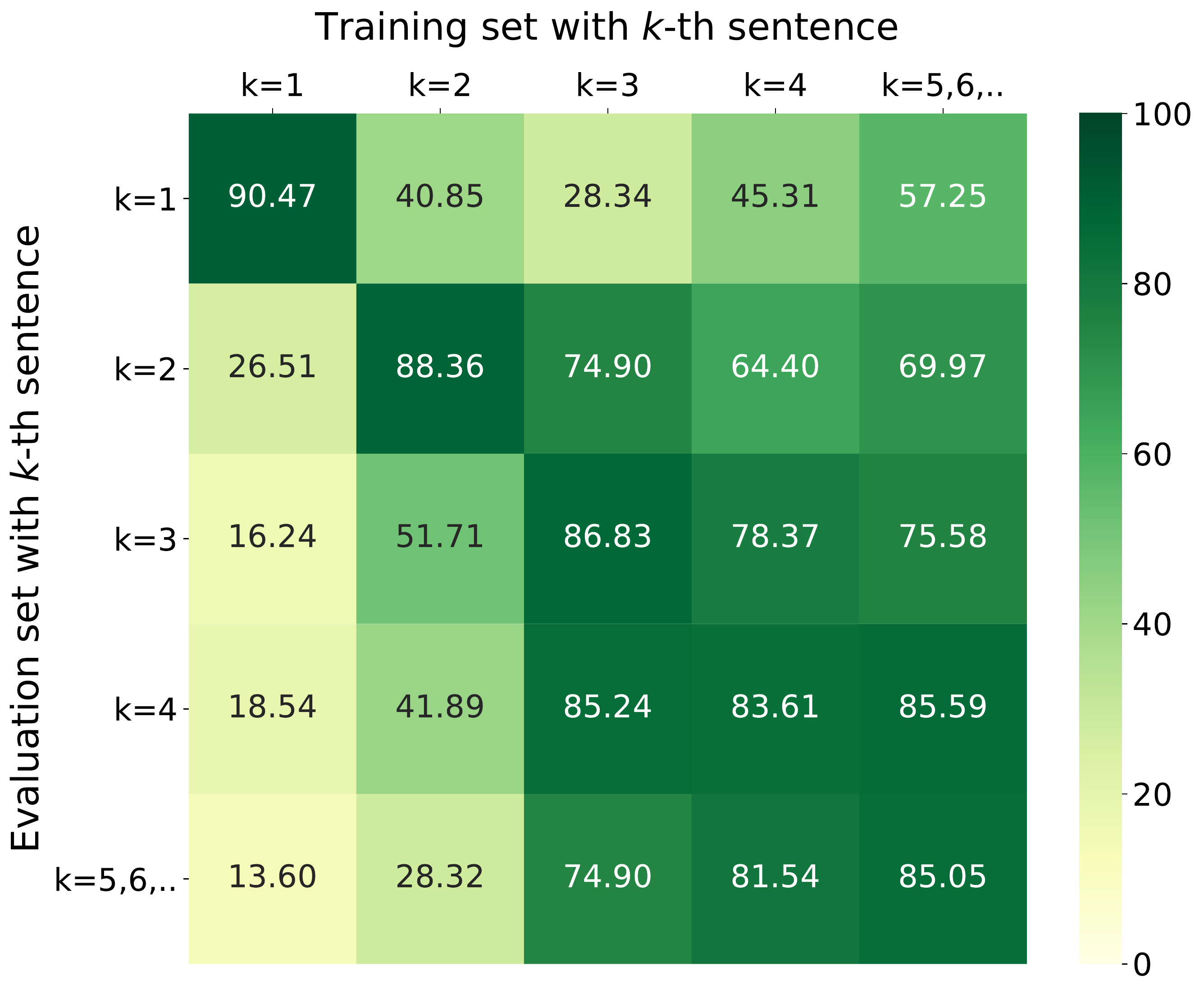}}
    \subfloat[BERT + Learned-Mixin]
     {\includegraphics[width=0.32\textwidth]{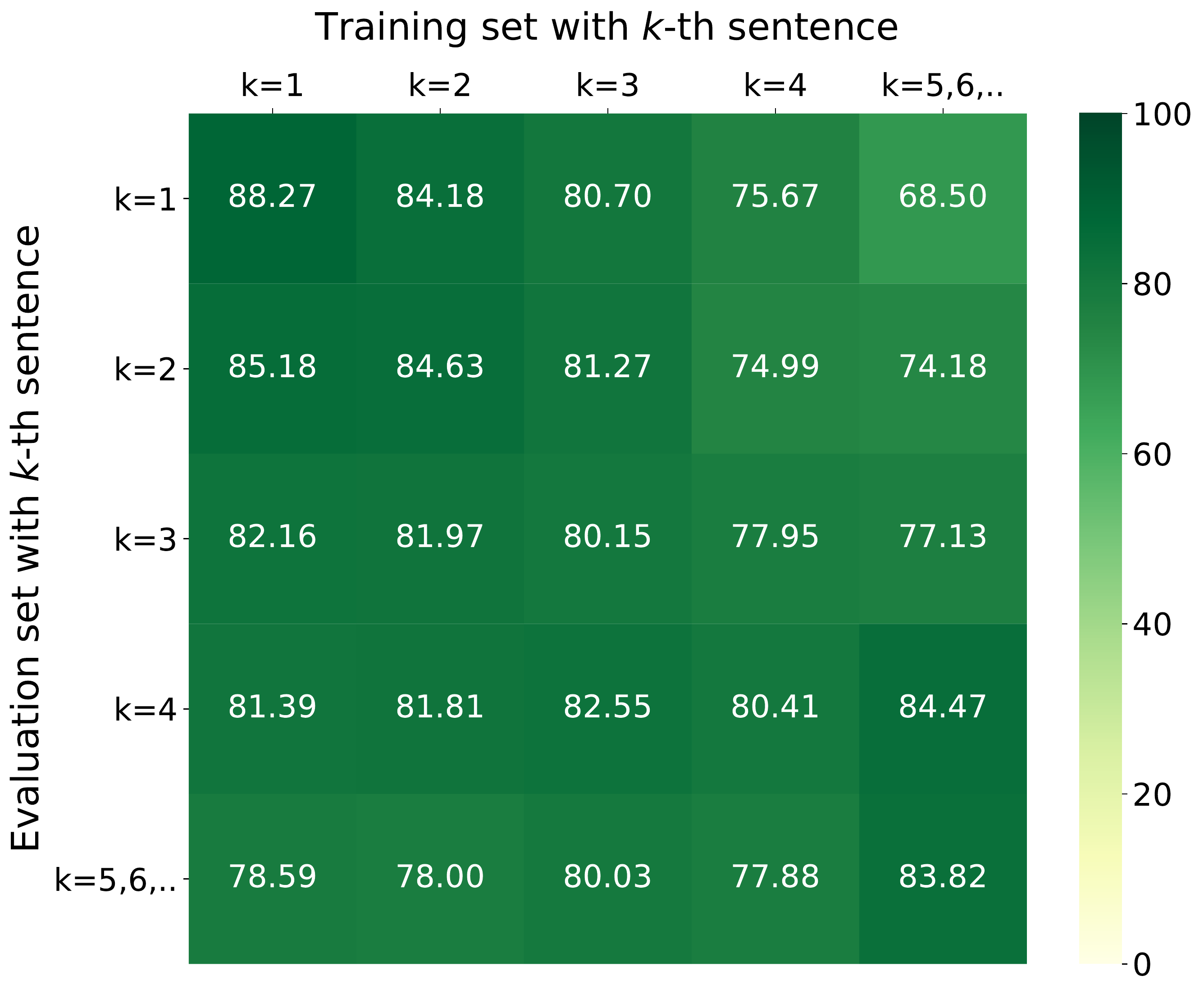}}
    \caption{
    Sentence-wise position bias in \squad. Models are trained on \squad\traink~and evaluated on \squad\validk.
    (a) Standard BERT suffers from position bias as the off-diagonal performance is significantly lower. 
    (b), (c) Our de-biasing method successfully handles the bias and provides consistently higher performance.}\vspace{-0.2cm}
    \label{fig:heatmap}
\end{figure*}

\paragraph{Results}
As shown in Table~\ref{other sentences}, all three models suffer from position bias in every sentence position while the learned-mixin method (+Learned-Mixin) successfully resolves the bias.
Due to the blurred sentence boundaries, position bias is less problematic when $k$ is large.
We observe a similar trend in BERT and XLNet while a huge performance drop is observed in BiDAF even with a large $k$. 

\paragraph{Visualization}
Figure~\ref{fig:heatmap} visualizes the sentence-wise position biases.
We train BERT, BERT + Bias Product and BERT + Learned Mixin on different subsets of SQuAD training set (\squad\traink) and evaluated on every \squad\validk~whose answers lie only in the $k$-th sentence.
As a result, the low performance in the off-diagonal represent the presence of position bias.
The figure shows that the biased model fails to predict the answers in different sentence positions (Figure~\ref{fig:heatmap} (a)) while our de-biased model achieves high performance regardless of the sentence position (Figure~\ref{fig:heatmap} (c)).
Again, as the value of $k$ increases, the boundary of the $k$-th sentence varies a lot in each sample, which makes the visualization of sentence-wise bias difficult.

\subsection{NewsQA and NaturalQuestions}
We test the effect of de-basing methods on datasets having different domains and different degrees of position bias.
NewsQA~\citep{trischler2017newsqa} is an extractive QA dataset that includes passages from CNN news articles.
NaturalQuestions~\citep{naturalquestions} is a dataset containing queries and passages collected from the Google search engine. 
We use the pre-processed dataset provided by the MRQA shared task~\citep{fisch-etal-2019-mrqa}.\footnote{\url{https://github.com/mrqa/MRQA-Shared-Task-2019}}

For each dataset, we construct two sub-training datasets; one contains samples with answers in the first sentence ($k=1$), and the other contains the remaining samples ($k=2,3,...$).
Models are trained on the original dataset and two sub-training datasets and evaluated on the original development set.

\paragraph{Implementation Details}
For NewsQA, we truncate each paragraph so that the length of each context is less than 300 words.
We eliminate training and development samples that become unanswerable due to the truncation.
For NaturalQuestions, we choose firstly occurring answers for training extractive QA models, which is a common approach in weakly supervised setting~\citep{joshi2017triviaqa, talmor2019multiqa}.

From NewsQA and NaturalQuestions, we construct two sub-training datasets having only the first annotated samples (\bigd\trainone) and the remaining samples (\bigd\trainafter). 
For a fair comparison, we fix the size of two sub-training sets to have 17,000 (NewsQA) and 40,000 samples (NaturalQuestions).

\begin{table} [t]
\centering
\resizebox{0.48\textwidth}{!}{
\def\arraystretch{0.9}
\begin{tabular}{l|ccc}
\toprule
\multirow{2}{*}{\textbf{NewsQA}\traink} & \multicolumn{3}{c}{\textbf{NewsQA}\valid} \\ 
& $\bm{k=\text{All}}$ & $\bm{k=1}$ & $\bm{k=2,3,...}$ \\\midrule
\textbf{BERT}            & \textbf{69.94} & 27.99 & 56.15 \\
+Bias Product            & 69.46 & 28.81 & 56.86  \\
+Learned-Mixin           & 69.42 & \textbf{44.50} & \textbf{58.22}  \\ 
\bottomrule
\end{tabular}}
\caption{F1 scores on NewsQA. Models are evaluated on the original development dataset (NewsQA\valid).}\vspace{-0.05cm}
\label{news}
\end{table}

\begin{table} [t]
\centering
\resizebox{0.48\textwidth}{!}{
\def\arraystretch{0.9}
\begin{tabular}{l|ccc}
\toprule
\multirow{2}{*}{\textbf{NQ}\traink} & \multicolumn{3}{c}{\textbf{NQ}\valid} \\
& $\bm{k=\text{All}}$ & $\bm{k=1}$  & $\bm{k=2,3,...}$\\\midrule
\textbf{BERT}            & 78.79 & 56.79 & 49.59 \\
+Bias Product            & 78.84 & 56.77 & 53.34  \\
+Learned-Mixin           & \textbf{79.04} & \textbf{72.83} & \textbf{60.63} \\ 
\bottomrule
\end{tabular}}
\caption{F1 scores on NaturalQuestions. Models are evaluated on the original development dataset (NQ\valid).}\vspace{-0.2cm}
\label{natural}
\end{table}

\paragraph{Results}
In Table~\ref{news} and Table~\ref{natural}, we show results of applying our methods.
In both datasets, BERT, trained on biased datasets ($k=1$ and $k=2,3,...$), significantly suffers from position bias.
Position bias is generally more problematic in the $k=1$ datasets while for NaturalQuestions, $k=2,3,...$ is also problematic.
Our de-biasing methods prevent performance drops in all cases without sacrificing the performance on the full training set ($k=\text{All}$).

\section{Related Work}
Various question answering datasets have been introduced with diverse challenges including reasoning over multiple sentences~\citep{joshi2017triviaqa}, answering multi-hop questions~\citep{yang2018hotpotqa}, and more~\citep{trischler2017newsqa,welbl2018constructing,naturalquestions,dua2019drop}.
Introduction of these datasets rapidly progressed the development of effective QA models~\citep{wang2016machine, seo2017bidirectional, xiong2017dcn+, wang2017gated, yu2018qanet, devlin2019bert, yang2019xlnet}, but most models predict the answer as positions without much discussion on it.

Our work builds on the analyses of dataset biases in machine learning models and ways to tackle them.
For instance, sentence classification models in natural language inference and argument reasoning comprehension suffer from word statistics bias~\citep{poliak2018hypothesis,minervini2018adversarially,kang2018adventure,belinkov132019adversarial, niven2019probing}. 
On visual question answering, models often ignore visual information due to the language prior bias~\citep{agrawal2016analyzing, zhang2016yin, goyal2017making, johnson2017clevr, agrawal2018don}. 
Several studies in QA also found that QA models do not leverages the full information in the given passage~\citep{chen2016thorough, min2018efficient, chen2019understanding, min2019compositional}.
Adversarial datasets have been also proposed to deal with this type of problem~\citep{jia2017adversarial, rajpurkar2018know}.
In this study, we define position bias coming from the prediction structure of QA models and show that positionally biased models can ignore information in different positions.

Our proposed methods are based on the bias ensemble method~\citep{clark2019don, he2019unlearn, mahabadi2020endtoend}.
Ensembling with the bias model encourages the model to solve tasks without converging to bias shortcuts.
\citet{clark2019don} conducted de-biasing experiments on various tasks including two QA tasks while they use tf-idf and the named entities as the bias models.

It is worth noting that several models incorporate the pointer network to predict the answer positions in QA~\citep{vinyals2015pointer, wang2016machine, wang2017gated}.
Also, instead of predicting positions, some models predict the n-grams as answers~\citep{lee2016learning, seo2019real}, generate answers in a vocabulary space~\citep{raffel2019exploring}, or use a generative model~\citep{lewisgenerative}.
We expect that these approaches suffer less from position bias and leave the evaluation of position bias in these models as our future work.

\section{Conclusion}
Most QA studies frequently utilize start and end positions of answers as training targets without much considerations.
Our study shows that most QA models fail to generalize over different positions when trained on datasets having answers in a specific position.
Our findings show that position can work as a spurious bias and alert researchers when building QA models and datasets.
We introduce several de-biasing methods to make models to ignore the spurious positional cues, and find out that the sentence-level answer prior is very useful.
Our findings also generalize to different positions and different datasets.
One limitation of our approach is that our method and analysis are based on a single paragraph setting which should be extended to a multiple paragraph setting to be more practically useful.

\section*{Acknowledgments}

This research was supported by a grant of the Korea Health Technology 
R\&D Project through the Korea Health Industry Development Institute (KHIDI), 
funded by the Ministry of Health \& Welfare, Republic of Korea (grant number: HR20C0021).
This research was also supported by National Research Foundation of Korea (NRF-2017R1A2A1A17069
645, NRF-2017M3C4A7065887).

\bibliography{anthology,emnlp2020}
\bibliographystyle{acl_natbib}

\clearpage

\numberwithin{figure}{section}
\numberwithin{table}{section}
\setcounter{page}{1}

\newpage
\appendix

\section{Implementation Details}\label{sec:train_detail}

\paragraph{Details of Training}
For all experiments, we use uncased BERT-base and cased XLNet-base.
We modify the open-sourced Pytorch implementation of models.\footnote{\url{ https://github.com/allenai/allennlp}, \url{https://github.com/huggingface/transformers}}
BiDAF is trained with the batch size of 64 for 30 epochs and BERT and XLNet are trained for 2 epochs with batch sizes 12 and 10, respectively.
The choice of hyperparameters mainly comes from the limitation of our computational resources and mostly follows the default setting used in their original works.
Note that our de-biasing methods do not require additional hyperparameters.

For all three models, the number of parameters remains the same as default settings with bias product and increases by a single linear layer with learned-mixin. 
We trained models on a single Titan X GPU.
The average training time of the bias ensemble method is similar to the original models.

\section{Examples of Answer Prior} \label{example_app}
\begin{figure*} [h]
    \centering
    \includegraphics[width=2\columnwidth]{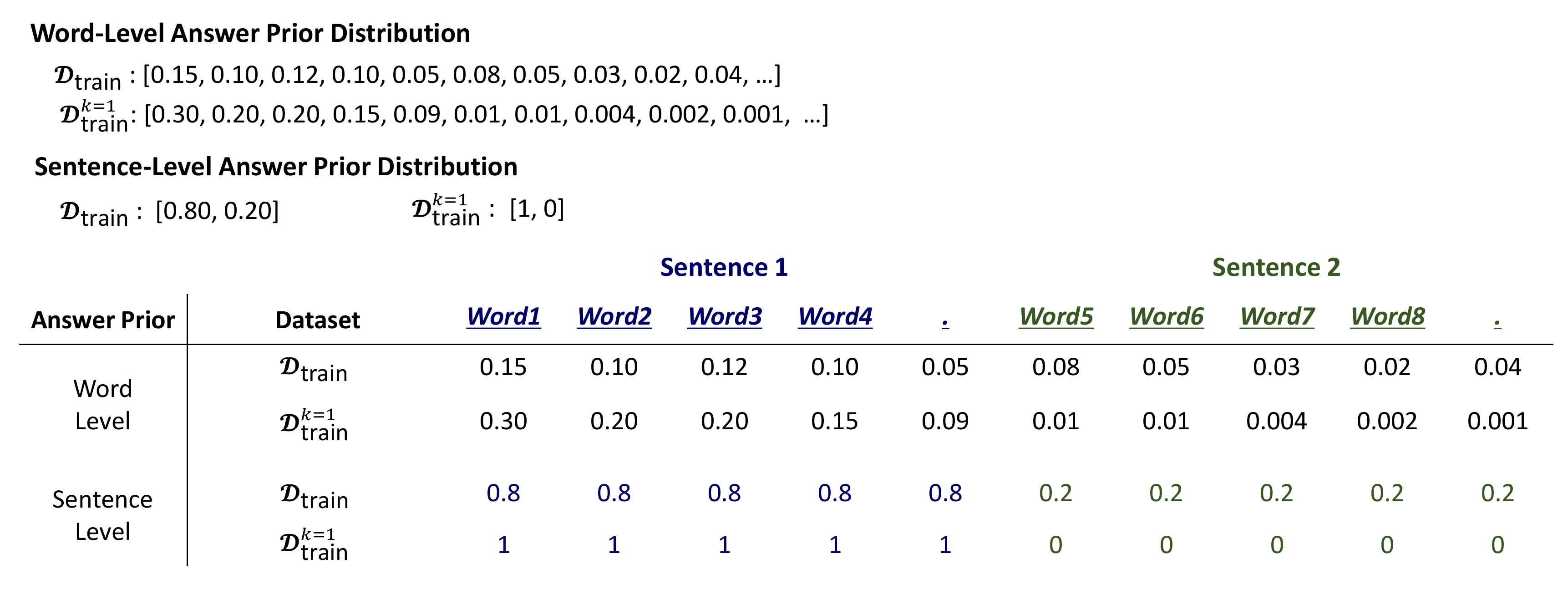}
    \caption{Example of three types of answer priors, word-level answer prior (Word-Level), sentence-level answer prior (Sentence-Level) and sentence-level answer prior on \bigd\trainone (Sentence-Level (First)).}\vspace{-0.5cm}
    \label{fig:example}
\end{figure*}

To provide a better understanding of our methods, Figure \ref{fig:example} shows examples of answer priors, which are used as bias models. See section~\ref{method} for detail.

\section{Visualization of Position Bias}
\begin{figure*} [h]
    \centering
    \includegraphics[width=1.1\columnwidth]{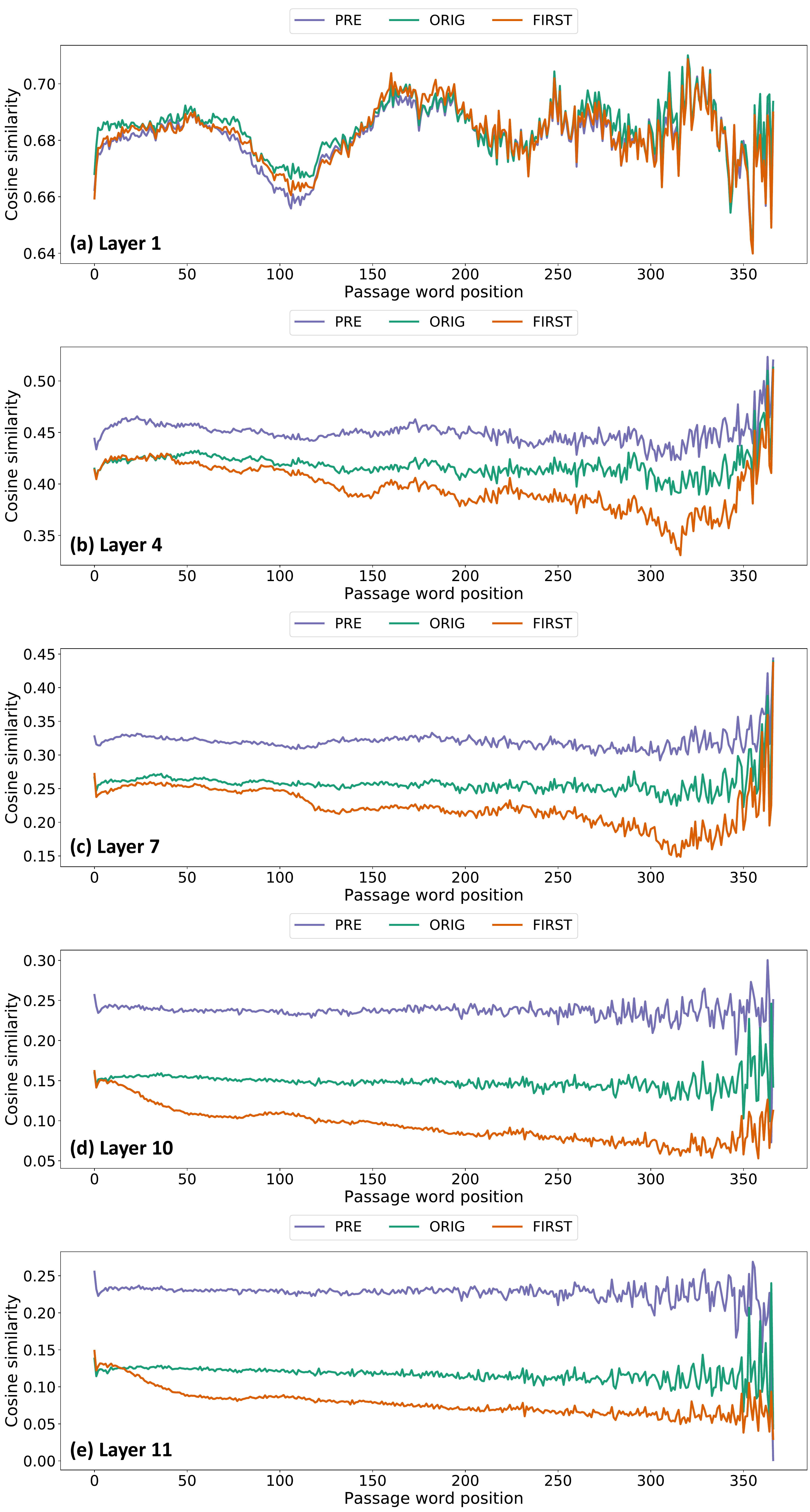}
    \caption{Visualization of each layer of BERT trained on \squad\train~ (\textcolor{green1}{\textsc{Orig}}), \squad\trainone~(\textcolor{orange}{\textsc{First}}), and BERT without fine-tuning (\textcolor{purple1}{\textsc{Pre}}). As the input passes each layer, position bias becomes more problematic.}\vspace{-0.5cm}
    \label{fig:app3}
\end{figure*}

\begin{figure*} [h]
    \centering
    \includegraphics[width=1.1\columnwidth]{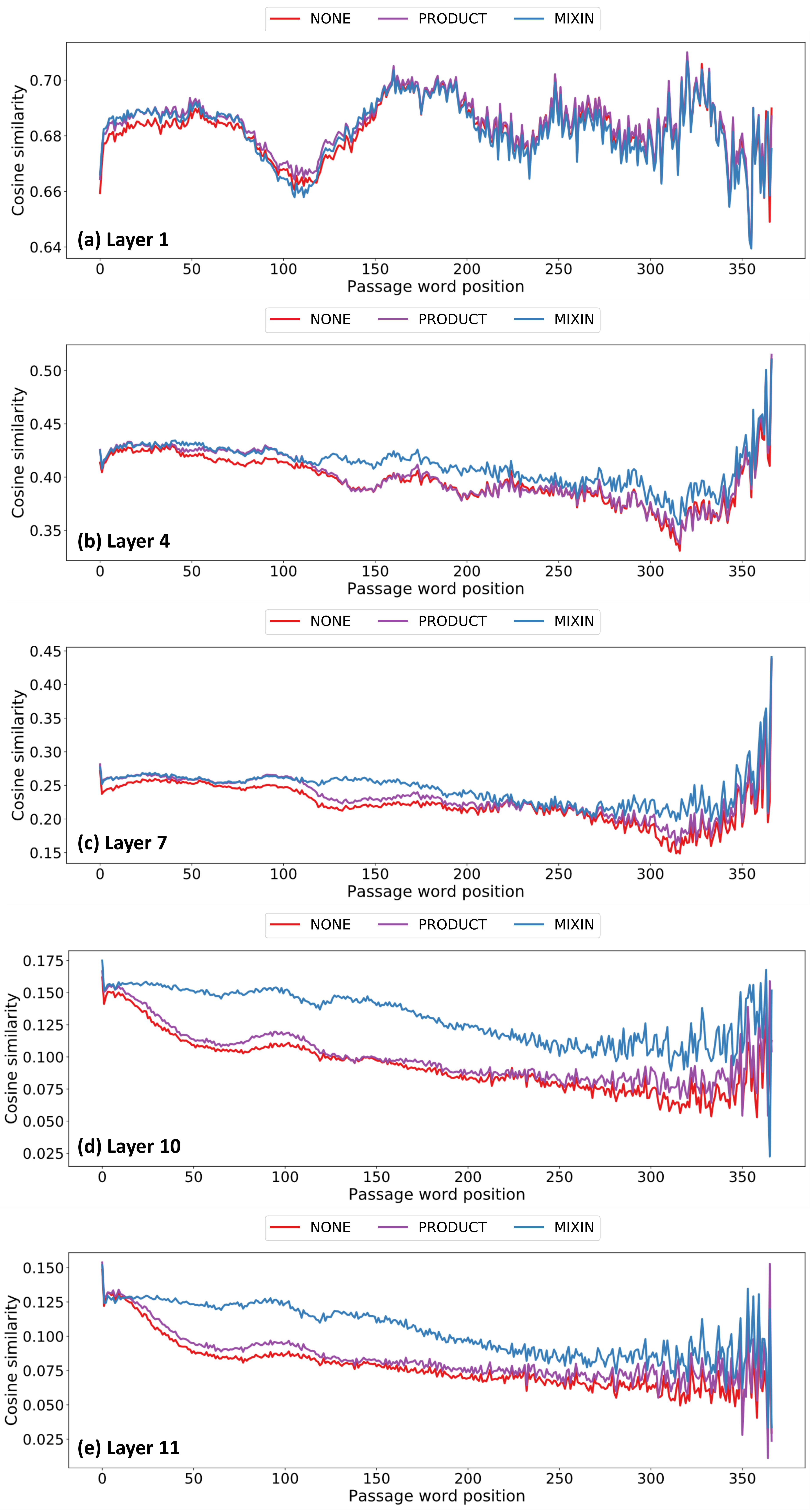}
    \caption{Visualization of each layer of de-biased BERT. BERT trained on \squad\trainone~without any de-biasing methods (\textcolor{red}{\textsc{None}}), with sentence-level prior bias product (\textcolor{purple2}{\textsc{Product}}), with learned-mixin (\textcolor{blue}{\textsc{Mixin}}). \textcolor{blue}{\textsc{Mixin}} preserves consistent information compared with \textcolor{red}{\textsc{None}} and prevents the bias propagation.}
    \label{fig:app4}
\end{figure*}

In Figure~\ref{fig:app3}, we plot the preserved amount of word information in the middle layers of BERT.
Figure~\ref{fig:app4} shows the effect of applying the de-biasing methods in each layer of BERT.
See Section~\ref{sec:2} and ~\ref{sec:4.1} for more detail.
We plot the results of layers 1, 4, 7, 10, and 11.

\end{document}

%% file: commands.tex
\newcommand{\squad}{SQuAD}

\newcommand{\bigd}{\textbf{$\bm{\mathcal{D}}$}}
\newcommand{\trainafter}{$^{k=2,3,...}_{\text{train}}$}
\newcommand{\trainone}{$^{k=1}_{\text{train}}$}
\newcommand{\traink}{$^{k}_{\text{train}}$}

\newcommand{\validafter}{$^{k=2,3,...}_{\text{dev}}$}
\newcommand{\validone}{$^{k=1}_{\text{dev}}$}
\newcommand{\validk}{$^{k}_{\text{dev}}$}
\newcommand{\train}{$_{\text{train}}$}
\newcommand{\valid}{$_{\text{dev}}$}

\definecolor{purple1}{HTML}{7570b3}
\definecolor{green1}{HTML}{1b9e77}
\definecolor{orange}{HTML}{d95f02}
\definecolor{green2}{HTML}{4daf4a}
\definecolor{red}{HTML}{e41a1c}
\definecolor{blue}{HTML}{377eb8}
\definecolor{purple2}{HTML}{984ea3}

